%% file: pan-notebook-template.tex
\documentclass[]{ceurart}

\usepackage{mathtools}
\usepackage{graphicx}
\usepackage{multirow}
\usepackage{adjustbox}
\usepackage[acronym]{glossaries}
\usepackage{wrapfig}
\usepackage[tableposition=top, labelfont=bf]{caption}
\usepackage[format=hang, justification=RaggedRight]{subcaption}
\usepackage{amssymb}
\usepackage[noend, linesnumbered, boxed]{algorithm2e}
\usepackage{floatrow}
\usepackage[skins]{tcolorbox}
\usepackage{mathtools}
\usepackage{amsmath}
\usepackage{bbm}
\usepackage{hyperref}
\newacronym{DML}{DML}{\textit{deep metric learning}}
\newacronym{LEV}{LEV}{\textit{linguistic embedding vector}}
\newacronym{BFS}{BFS}{\textit{Bayes factor scoring}}
\newacronym{UAL}{UAL}{\textit{uncertainty adaptation layer}}
\newacronym{AV}{AV}{authorship verification}
\newacronym{LEVs}{LEVs}{\textit{linguistic embedding vectors}}
\newacronym{O2D2}{O2D2}{\textit{out-of-distribution detection}}
\newacronym{ECE}{\texttt{ECE}}{\textit{expected calibration error}}
\newacronym{MCE}{\texttt{MCE}}{\textit{maximum calibration error}}
\newacronym{PLDA}{PLDA}{\textit{probabilistic linear discriminant analysis}}
\newacronym{acc}{\texttt{acc}}{\textit{accuracy}}
\newacronym{conf}{\texttt{conf}}{\textit{confidence score}}

\newcommand{\loss}{\ensuremath{\mathcal{L}}}

\begin{document}

\copyrightyear{2021}
\copyrightclause{Copyright for this paper by its authors.
  Use permitted under Creative Commons License Attribution 4.0
  International (CC BY 4.0).}

\conference{CLEF 2021 -- Conference and Labs of the Evaluation Forum, 
	September 21--24, 2021, Bucharest, Romania}

\title{O2D2: Out-Of-Distribution Detector to Capture Undecidable Trials in Authorship Verification}
\title[mode=sub]{Notebook for PAN at CLEF 2021}

\author[1]{Benedikt Boenninghoff}[%
email=benedikt.boenninghoff@rub.de,
]
\address[1]{Ruhr University Bochum, Germay}

\author[2]{Robert M. Nickel}[%
email=rmn009@bucknell.edu,
]
\address[2]{Bucknell University, USA}

\author[1]{Dorothea Kolossa}[%
email=dorothea.kolossa@rub.de,
]

\begin{abstract}
The PAN 2021 authorship verification (AV) challenge is part of a three-year strategy, moving from a cross-topic/closed-set AV task to a cross-topic/open-set AV task over a collection of fanfiction texts. In this work, we present a novel hybrid neural-probabilistic framework that is designed to tackle the challenges of the 2021 task. Our system is based on our 2020 winning submission, with updates to significantly reduce sensitivities to topical variations and to further improve the system’s calibration by means of an uncertainty adaptation layer. Our framework additionally includes an \textit{out-of-distribution} detector (O2D2) for defining non-responses. Our proposed system outperformed all other systems that participated in the PAN 2021 AV task. 
\end{abstract}


\maketitle

\setlength{\textfloatsep}{8pt}
\setlength{\abovedisplayskip}{5.pt}
\setlength{\belowdisplayskip}{6.pt}
\setlength{\abovedisplayshortskip}{5.pt}
\setlength{\belowdisplayshortskip}{6.pt}

\input{sections/introduction}
\input{sections/preprocessing}
\input{sections/model}
\input{sections/experiments}
\input{sections/conclusion}

\vspace*{-.3cm}
\begin{acknowledgments}
\vspace*{-.2cm}
This work was in significant parts performed on an HPC cluster at Bucknell University through the support of the National Science Foundation, Grant Number 1659397. Project funding was provided by the state of North Rhine-Westphalia within the Research Training Group "SecHuman - Security for Humans in Cyberspace" and by the Deutsche Forschungsgemeinschaft (DFG) under Germany’s Excellence Strategy - EXC2092CaSa- 390781972. 
\end{acknowledgments}

\vspace*{-0.3cm}
\small
\bibliography{references/refs}

\end{document}

%% file: sections/introduction.tex
\vspace*{-.8cm}
\section{Introduction}
\vspace*{-.2cm}

In this paper we are proposing a significant extension to the authorship verification (AV) system presented in~\cite{boenninghoff:2020a}. The work is part of the PAN 2021 AV shared task\cite{bevendorff:2021b}, for which the PAN organizers provided the challenge participants with a publicly available dataset of fanfiction.

Fanfiction texts are fan-written extensions of well-known story lines, in which the so-called fandom topic describes the principal subject of the literary document (e.g. \emph{Harry Potter}). The use of fanfiction as a {\em genre\/} has three major advantages. Firstly, the abundance of texts written in this genre makes it feasible to collect a large training dataset and, therefore, to build more complex \gls{AV} systems based on modern deep learning techniques, which will hopefully boost progress in this research area. Additionally, fanfictional documents also come with meaningful meta-data like topical information, which can be used to investigate the topical interference in authorship analysis. Lastly, although the documents are usually produced by non-professional writers, contrary to social media messages, they usually follow standard grammatical and spelling conventions. This allows participants to incorporate pretrained models for, e.g., part-of-speech tagging, and to reliably extract traditional stylometric features~\cite{Stamatatos2009}.

The previous edition of the PAN \gls{AV} task dealt with cross-fandom/closed-set \gls{AV}~\cite{kestemont:2020}. 
The objective of the \textit{cross-fandom} \gls{AV} task is to automatically decide whether two fanfictional documents covering different fandoms belong to the same author.
The term \textit{closed-set} refers to the fact that the test dataset, which is not publicly available, only contains trials from a subset of the authors and fandoms provided in the training data.

To increase the level of difficulty, the current PAN AV challenge moved from a \textit{closed-set} task to an \textit{open-set} task in 2021, while the training dataset is identical to that of the previous year~\cite{kestemont:2021}.
In this scenario, the new test data contains \textit{only} authors and fandoms that were \textit{not} included in the training data. We thus expect a \textit{covariate shift} between training and testing data, i.e.~the distribution of our neural stylometric representations extracted from the training data is expected to be different from the distribution of the test data representations.
It was implicitly shown in~\cite{kestemont:2020}, and our experiments confirm this analysis, that such a covariate shift, due to topic variability, is a major cause of errors.

\vspace*{-.4cm}
\section{System Overview}
\begin{figure}[t]
\centering
\includegraphics[width=1.0\textwidth]{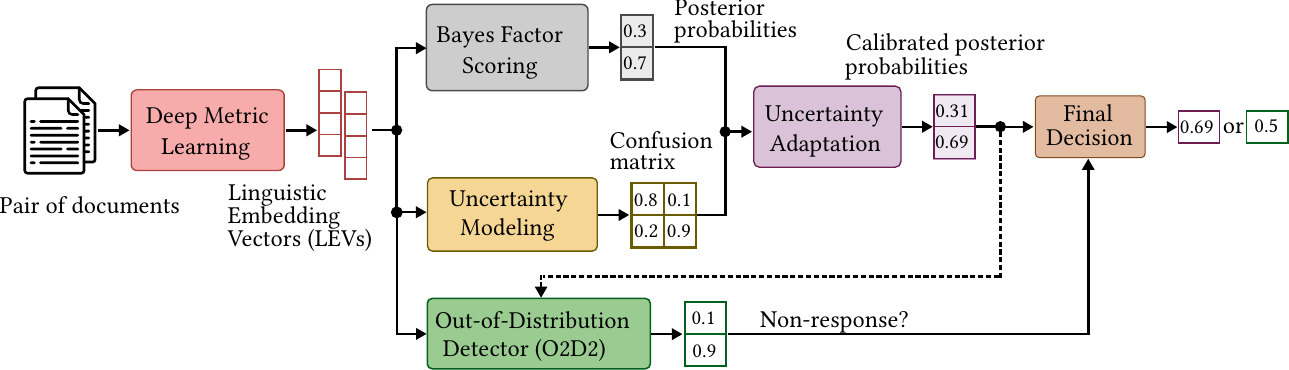}
\caption{Our proposed hybrid neural-probabilistic framework for the PAN 2021 cross-fandom open-set authorship verification task.} 
\label{fig:model21}
\vspace{-0.2cm}
\end{figure}

\vspace*{-.2cm}
The overall structure of our revised system\footnote{The source code is accessible online: \url{https://github.com/boenninghoff/pan\_2020\_2021\_authorship\_verification}} is shown in~Fig.~\ref{fig:model21}. It expands our winning system from 2020 as follows:
Suppose we have a pair of documents $\mathcal{D}_1$ and $\mathcal{D}_2$ with an associated ground-truth hypothesis $\mathcal{H}_a$ for $a\in\{0,1\}$.
 The value of $a$ indicates, whether the two documents were written by the same author ($a=1$) or by different authors ($a=0$).
 Our task can formally be expressed as a mapping
$f\!\!:\!\!\{\mathcal{D}_1, \mathcal{D}_2\} \longrightarrow
p \in [0,1]$. 
The estimated label $\widehat{a}$ is obtained from a threshold test applied to the output prediction $p$. In our case, we choose $\widehat{a}=1$ if $p>0.5$ and $\widehat{a}=0$ if $p<0.5$. The PAN 2020/21 shared tasks also permit the return of a \textit{non-response} (in addition to $\widehat{a}=1$ and $\widehat{a}=0$)
in cases of high uncertainty~\cite{kestemont:2020}, e.g.~when $p$ is close to 0.5. In this work, we therefore define three hypotheses:   
\begin{align*}
        \mathcal{H}_0:~~&\text{The two documents were written by two different persons,} \\
        \mathcal{H}_1:~~&\text{The two documents were written by the same person,} \\
    \mathcal{H}_2:~~&\text{Undecidable, trial does not suffice to establish authorship.}
\end{align*}

In~\cite{boenninghoff:2020a}, we introduced the concept of \gls{LEVs}. To obtain these, we perform neural feature extraction followed by \gls{DML} to encode the stylistic characteristics of a pair of documents into a pair of fixed-length and topic-invariant stylometric representations.
Given the \gls{LEVs}, a \gls{BFS} layer computes the posterior probability for a trial. This discriminative two-covariance model was introduced in~\cite{6466371}.
As a new component, we propose an \gls{UAL}.
This idea is adopted from~\cite{luo-etal-2017-learning-noise}, aiming to find and correct wrongly classified trials of the \gls{BFS} layer, to model its noise behavior, and to return re-calibrated posteriors.

For the decision whether to accept $\mathcal{H}_0$/$\mathcal{H}_1$, or to return a non-response, i.e.~$\mathcal{H}_2$, it is desirable that the value of the posterior $p$ reliably reflects the uncertainty of the decision-making process.
We may roughly distinguish two different types of uncertainty~\cite{NIPS2017_2650d608}: In \gls{AV}, \textit{aleatoric} or data uncertainty is associated with properties of the document pairs.
Examples are topical variations or the intra- and inter-author variabilities. 
Aleatoric uncertainty generally can not be reduced, but it can be addressed (to a certain extent) by returning a non-response (i.e. hypothesis $\mathcal{H}_2$) if it is too large to allow for a reliable decision. 
To accomplish this, and inspired by~\cite{Shao2020CalibratingDN}, we incorporate a feed-forward network for \gls{O2D2}, which is trained on a dataset that is different, i.e.~disjoint w.r.t.~authors and fandoms, from the training set used to optimize the \gls{DML}, \gls{BFS} and \gls{UAL} components. 

Additionally, \textit{epistemic} or model uncertainty characterizes uncertainty in the model parameters. Examples are \textit{unseen} authors or topics.
Epistemic uncertainty can be
reduced through a substantial increase in the amount of training data, i.e.~an increase in the number of training pairs.
We capture epistemic uncertainty in our work through the proposed \gls{O2D2} approach and also by extending our model to an ensemble.
We expect all models to behave similarly for known authors or topics, but the output predictions may be widely dispersed for pairs under covariate shift~\cite{10.5555/3295222.3295387}.


The training procedure consists of two stages: In the first stage, we simultaneously train the \gls{DML}, \gls{BFS} and \gls{UAL} components. In the second stage, we learn the parameters of the \gls{O2D2} model.

%% file: sections/preprocessing.tex
\vspace*{-.4cm}
\section{Dataset Splits for the PAN 2021 AV Task}
\vspace*{-.2cm}

The text preprocessing strategies, including tokenization and pair re-sampling, are comprehensively described in~\cite{boenninghoff:2021}. The \textit{fanfictional} dataset for the PAN 2020/21 \gls{AV} tasks are described in~\cite{kestemont:2020, kestemont:2021}. In the following, we report on the various dataset splits that we employed for our PAN 2021 submission.

Each document pair is characterized by a tuple $(a, f)$, where $a\in \{0, 1\}$ denotes the {\em authorship similarity label\/} and $f\in \{0, 1\}$ describes the equivalent for the fandom. We assign each document pair to one of the following author-fandom  subsets\footnote{\texttt{SA}=same author, \texttt{DA}=different authors,
\texttt{SF}=same fandom, \texttt{DF}=different fandoms}
\texttt{SA\_SF}, \texttt{SA\_DF}, \texttt{DA\_SF},  and \texttt{DA\_DF}
given its label tuple $(a,f)$.

\begin{figure}[t]
\centering
\includegraphics[width=.9\textwidth]{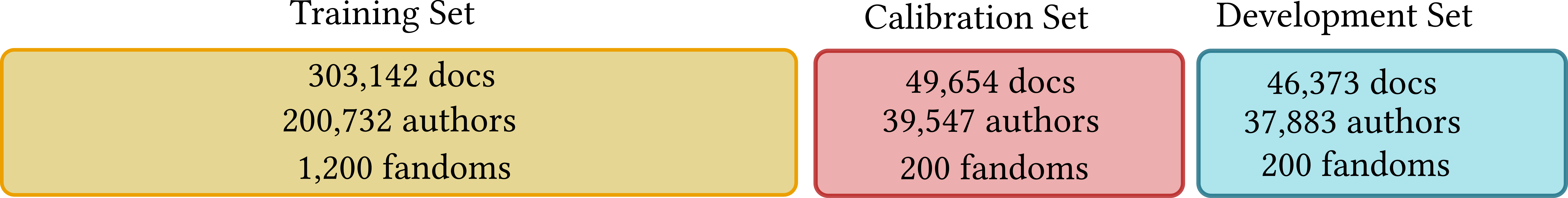}
\vspace{-0.05cm}
\caption{Disjoint splits of the large PAN 2020/21 training set.}
\label{fig:datasplit}
\vspace{-0.2cm}
\end{figure}

As shown in~\cite{boenninghoff:2021},
one of the difficulties working with the provided small/large PAN datasets is that each  author  generally contributes only with a small number of documents. As a result, we observe a high degree of overlap in the re-sampled subsets of same-author trials.
We decided to work only with the large dataset this year and split the documents into three disjoint (w.r.t.~authorship and fandom) sets.
Overlapping  documents, where author and fandom belong to different sets, are removed. The splits are summarized in Fig.~\ref{fig:datasplit} and Table~\ref{tab:pairresampling}. Altogether, the following datasets have been involved in the PAN 2021 shared task, to train the model components, tune the hyper-parameter and for testing:
\vspace*{-0.11cm}
\begin{itemize}
    \item The \textbf{training set} is identical to the one used in~\cite{boenninghoff:2021} and was employed for the first stage, i.e., to train the \gls{DML}, \gls{BFS} and \gls{UAL} components simultaneously. During training we re-sampled the pairs epoch-wise such that all documents contribute equally to the neural network training in each epoch. The numbers of training pairs provided in Table~\ref{tab:pairresampling} therefore vary in each epoch.
    \item The \textbf{calibration set} has been used for the second stage, i.e., to train (calibrate) the \gls{O2D2} model. During training, we again re-sampled the pairs in each epoch and limited the total number of pairs in the different-authors subsets to partly balance the dataset.
    \item The purpose of the \textbf{validation set} is to tune the hyper-parameters of the \gls{O2D2} stage and to report the final evaluation metrics for 
    all stages in Section~\ref{sec:exp}.  
    \item The \textbf{development set} is identical to the evaluation set in\cite{boenninghoff:2021} and was used to tune the hyper-parameters during the training of the first stage. This dataset contains pairs from the calibration and validation sets. However, due to the pair re-sampling strategy in~\cite{boenninghoff:2021}, documents may appear in different subsets and varied document pairs may be sampled. It thus does not represent a union of the calibration and validation sets.
    \item Finally, the \textbf{PAN 2021 evaluation set}, which is not publicly available, has been used to test our submission and to compare it with the proposed frameworks of all other participants. 
\end{itemize}
\vspace*{-0.1cm}
Note that both, the validation and development set in Table~\ref{tab:pairresampling} only contain \texttt{SA\_DF} and \texttt{DA\_SF} pairs, for reasons discussed in Section~\ref{sec:exp}. 
The pairs of these sets are sampled once and then kept fixed.

 \begin{table}[t]
\caption{Numbers of (re-)sampled pairs for all datasets.}
\vspace{-0.1cm}
\centering
\resizebox{0.56\textwidth}{!}{
  \begin{tabular}{|c |c | c |c |c |}
    \hline
 Dataset & \texttt{SA\_SF}
 &\texttt{SA\_DF}           &\texttt{DA\_SF} &\texttt{DA\_DF}    
 \\ \hline
 Training set &16,045 &28,500      &64,300    &42,730 \\
 Calibration set &2,100      &2,715    &4,075 &4,075 \\
Validation set &0 &2,280      &3075    &0 \\
 Development set &0 &5,215      &7,040    &0
    \\ \hline
\end{tabular}
}
\label{tab:pairresampling}
\end{table}

%% file: sections/model.tex
\vspace*{-.4cm}
\section{Methodologies}
\vspace*{-.2cm}
\label{sec:NPM}
In this section, we briefly describe all components of our neural-probabilistic model. Sections~\ref{seq:neural_dml} through~\ref{CombLossFun} repeat information that is already provided in~\cite{boenninghoff:2021} to provide proper context.

\vspace*{-.2cm}
\subsection{Neural Feature Extraction and Deep Metric Learning}
\vspace*{-.1cm}
\label{seq:neural_dml}
Feature extraction and deep metric learning are realized in the form of a \textit{Siamese} network, feeding both input documents through exactly the same function.

\vspace*{-.2cm}
\subsubsection{Neural Feature Extraction: }
\vspace*{-.1cm}
The system passes token and character embeddings into a two-tiered bidirectional LSTM network with attentions, 
\begin{align}
    \label{eq:docemb}
    \boldsymbol{x}_i = \text{NeuralFeatureExtraction}_{\boldsymbol{\theta}}\big(
        \boldsymbol{E}_i^w, 
        \boldsymbol{E}_i^c
    \big),
\end{align}
where $\boldsymbol{\theta}$ contains all trainable parameters, 
$\boldsymbol{E}^w_i$ represents word embeddings and $\boldsymbol{E}^c_i$ represents character embeddings.
A comprehensive description is given in~\cite{boenninghoff:2019b}.

\vspace*{-.2cm}
\subsubsection{Deep Metric Learning: }
\vspace*{-.1cm}
We feed the document embeddings $\boldsymbol{x}_i$ in Eq.~\eqref{eq:docemb} into a metric learning layer,
$
\boldsymbol{y}_i = \tanh\big(\boldsymbol{W}^{\textsc{DML}}  \boldsymbol{x}_i +  \boldsymbol{b}^{\textsc{DML}}\big),
$
which yields the two \gls{LEVs} $\boldsymbol{y}_1$ and $\boldsymbol{y}_2$ via the trainable parameters $\boldsymbol{\psi}=\{\boldsymbol{W}^{\textsc{DML}},$ $\boldsymbol{b}^{\textsc{DML}}\}$. 
We then compute the Euclidean distance between both LEVs,
$
    d(\boldsymbol{y}_1,\boldsymbol{y}_2) = \left\lVert \boldsymbol{y}_1 - \boldsymbol{y}_2 \right\rVert_2^2.
$
In~\cite{boenninghoff:2021}, we introduced a new \textit{probabilistic} version of the contrastive loss: Given the Euclidean distance of the \gls{LEVs}, we apply a kernel function \begin{align}
\label{eq:DMLl}
p_{\text{DML}}(\mathcal{H}_1 |\boldsymbol{y}_1,\boldsymbol{y}_2) = \exp\big(- \gamma~ d(\boldsymbol{y}_1,\boldsymbol{y}_2)^{\alpha} \big),
\end{align}
where $\gamma$ and $\alpha$ can be seen as both, hyper-parameters or trainable variables. The loss then is given by

\vspace{-0.2in}

\begin{align}
 \begin{split}
 \label{eq:lossdml2}
 \loss^{\text{DML}}_{\boldsymbol{\theta}, \boldsymbol{\psi}} = a \cdot \max \big\{ \tau_s- p_{\text{DML}}(\mathcal{H}_1 |\boldsymbol{y}_1,\boldsymbol{y}_2), 0 \big\}^2
 +       (1-a) \cdot \max \left\{ p_{\text{DML}}(\mathcal{H}_1 |\boldsymbol{y}_1,\boldsymbol{y}_2) - \tau_d, 0\right\}^2,
\end{split}
\vspace{-0.4in}
\end{align}
where we set $\tau_s = 0.91$ and $\tau_d=0.09$.

\vspace*{-.2cm}
\subsection{Deep Bayes Factor Scoring}
\vspace*{-.1cm}
We assume that the \gls{LEVs} stem from a Gaussian generative model that can be decomposed as $\boldsymbol{y} = \boldsymbol{s} + \boldsymbol{n}$,
where $\boldsymbol{n}$ characterizes a noise term.
We assume that the writing characteristics of the author lie in a latent stylistic variable $\boldsymbol{s}$.
The probability density functions for $\boldsymbol{s}$ and $\boldsymbol{n}$ are modeled as Gaussian distributions.
We outlined in~\cite{boenninghoff:2020a} how to compute 
the likelihoods for both hypotheses.
The verification score for a trial is then given by the log-likelihood ratio: $ \text{score}(\boldsymbol{y}_1, \boldsymbol{y}_2) 
 = \log p(\boldsymbol{y}_1, \boldsymbol{y}_2 |\mathcal{H}_1) - \log p(\boldsymbol{y}_1, \boldsymbol{y}_2 |\mathcal{H}_0)$.
Assuming $p(\mathcal{H}_1) = p(\mathcal{H}_0) = \frac{1}{2}$, the  probability for a same-author trial is calculated as~\cite{boenninghoff:2020a}:
\begin{align}
 \label{eq:llrs}
  p_{\text{BFS}}(\mathcal{H}_1|\boldsymbol{y}_1, \boldsymbol{y}_2)
        = \frac{p(\boldsymbol{y}_1, \boldsymbol{y}_2|\mathcal{H}_1)}
            {p(\boldsymbol{y}_1, \boldsymbol{y}_2|\mathcal{H}_1) + p(\boldsymbol{y}_1, \boldsymbol{y}_2|\mathcal{H}_0)}
        = \text{Sigmoid}\big( \text{score}(\boldsymbol{y}_1, \boldsymbol{y}_2) \big)
\end{align}
We reduce the dimension of the \gls{LEVs} via
$
    \boldsymbol{y}_i^{\textsc{BFS}} = \text{tanh}\big(\boldsymbol{W}^{\textsc{BFS}} \boldsymbol{y}_i +  \boldsymbol{b}^{\textsc{BFS}}\big)
$
to ensure numerically stable inversions of the matrices~\cite{boenninghoff:2020a}.
We rewrite Eq.~\eqref{eq:llrs} as
\begin{align}
 \label{eq:scorefinal} 
  p_{\text{BFS}}(\mathcal{H}_1|\boldsymbol{y}_1, \boldsymbol{y}_2)
        = \text{Sigmoid}\big( \text{score}(\boldsymbol{y}_1^{\textsc{BFS}}, \boldsymbol{y}_2^{\textsc{BFS}}) \big)
\end{align}
and incorporate Eq.~\eqref{eq:scorefinal} into the binary cross entropy,
\begin{align}
\label{eq:loss2}
\loss^{\text{BFS}}_{\boldsymbol{\phi}} 
    = a \cdot \log \left\{ p_{\text{BFS}}(\mathcal{H}_1|\boldsymbol{y}_1, \boldsymbol{y}_2) \right\}
            + (1-a) \cdot \log \left\{ 1 - p_{\text{BFS}}(\mathcal{H}_1|\boldsymbol{y}_1, \boldsymbol{y}_2) \right\},
\end{align}
where all trainable parameters are denoted with $\boldsymbol{\phi} = \big\{\boldsymbol{W}^{\textsc{BFS}}, \boldsymbol{b}^{\textsc{BFS}}, \boldsymbol{W}, \boldsymbol{B}, \boldsymbol{\mu} \big\}$.

\vspace*{-.2cm}
\subsection{Uncertainty Modeling and Adaptation}
\vspace*{-.1cm}
Now, we treat the posteriors of the \gls{BFS} component as noisy outcomes and rewrite Eq.~\eqref{eq:scorefinal} as
$p_{\text{BFS}}(\widehat{\mathcal{H}}_1|\boldsymbol{y}_1, \boldsymbol{y}_2)$
to emphasize that this represents an estimated posterior.
We firstly have to find a single representation for both \gls{LEVs}, which is done by
$
    \boldsymbol{y}^{\textsc{UAL}} = \tanh\big(\boldsymbol{W}^{\textsc{UAL}} 
    \big( \boldsymbol{y}_1 - \boldsymbol{y}_2 \big)^{\circ 2} +  \boldsymbol{b}^{\textsc{UAL}}\big),
$
where 
$(\cdot)^{\circ 2}$ denotes the element-wise square. 
Next, we compute a $2\times 2$ confusion matrix as follows
\begin{align}
    \label{eq:ualC}
    p(\mathcal{H}_j | \widehat{\mathcal{H}}_i, \boldsymbol{y}_1, \boldsymbol{y}_2) =  \frac{\exp\big( \boldsymbol{w}_{ji}^T ~\boldsymbol{y}^{\textsc{BFS}}  + b_{ji} \big)}
{\sum\limits_{i'\in\{0,1\}}  \exp\big(\boldsymbol{w}_{ji'}^T ~ \boldsymbol{y}^{\textsc{BFS}} + b_{ji'} \big)}
\quad \text{ for } i,j\in \{0,1\}.
\end{align}
The term $ p(\mathcal{H}_j | \widehat{\mathcal{H}}_i, \boldsymbol{y}_1, \boldsymbol{y}_2)$ defines the conditional probability of the true hypothesis $\mathcal{H}_j$ given 
the hypothesis $\widehat{\mathcal{H}}_i$ assigned by the \gls{BFS}.
We can then define the final output predictions as:
\begin{align}
\label{eq:UAL}
p_{\text{UAL}}(\mathcal{H}_j| \boldsymbol{y}_1, \boldsymbol{y}_2)
 = \sum\limits_{i\in \{0,1\}}    
        p(\mathcal{H}_j | \widehat{\mathcal{H}}_i, \boldsymbol{y}_1, \boldsymbol{y}_2)
        \cdot p_{\text{BFS}}(\widehat{\mathcal{H}}_i | \boldsymbol{y}_1, \boldsymbol{y}_2).
\end{align}
The loss consists of two terms, the negative log-likelihood of the ground-truth hypothesis and a regularization term,
\begin{align}
\label{eq:lossual}
 \begin{split}
\loss^{\text{UAL}}_{\boldsymbol{\lambda}} 
   = -\log p_{\text{UAL}}(\mathcal{H}_j| \boldsymbol{y}_1, \boldsymbol{y}_2) 
   + \beta~ \sum_{i\in\{0,1\}} \sum_{j\in \{0,1\}}
    p(\mathcal{H}_j | \widehat{\mathcal{H}}_i, \boldsymbol{y}_1, \boldsymbol{y}_2) 
    \cdot
    \log p(\mathcal{H}_j | \widehat{\mathcal{H}}_i, \boldsymbol{y}_1, \boldsymbol{y}_2),
\end{split}
\end{align}
with trainable parameters denoted by $\boldsymbol{\lambda} = \big\{\boldsymbol{W}^{\textsc{UAL}}, \boldsymbol{b}f^{\textsc{UAL}}, \boldsymbol{w}_{ji}, \boldsymbol{b}_{ji} | j,i\in \{0,1\}\big\}$.
The regularization term, controlled by $\beta$, follows the maximum entropy principle to penalize the confusion matrix for returning over-confident posteriors~\cite{pereyra2017regularizing}. 

\vspace*{-.2cm}
\subsection{Combined Loss Function:}
\vspace*{-.1cm}
\label{CombLossFun}
All components are optimized independently w.r.t.~the following combined loss:
\begin{align}
\label{eq:lossall}
 \begin{split}
\loss_{\boldsymbol{\theta}, \boldsymbol{\psi},\boldsymbol{\phi},  \boldsymbol{\lambda}} 
 = \loss^{\text{DML}}_{\boldsymbol{\theta}, \boldsymbol{\psi}} 
+ \loss^{\text{BFS}}_{\boldsymbol{\phi}} 
 + \loss^{\text{UAL}}_{\boldsymbol{\lambda}}. 
\end{split}
\end{align}

\subsection{Out-of-Distribution Detector (O2D2)}
Following~\cite{Shao2020CalibratingDN}, we
incorporate a second neural network to detect 
undecidable trials. 
We treat the training procedure as a binary verification task.
Given the learned \gls{DML}, \gls{BFS} and \gls{UAL} components,
the estimated authorship labels are obtained via
\begin{align}
    \widehat{a} = \arg\max \big[p_{\text{UAL}}(\mathcal{H}_0| \boldsymbol{y}_1, \boldsymbol{y}_2) , ~p_{\text{UAL}}(\mathcal{H}_1| \boldsymbol{y}_1, \boldsymbol{y}_2)\big].
\end{align}
Now, we can define the binary \gls{O2D2} labels as follows:
\begin{equation}
\label{eqn:oodd_decision}
l^{\text{O2D2}} = 
\begin{cases}
1,  & \text{if } a \ne \widehat{a}
      \text{~~~or~~~} 0.5 - \epsilon \le p_{\text{UAL}}(\mathcal{H}_1| \boldsymbol{y}_1, \boldsymbol{y}_2) \le 0.5 + \epsilon
,\\ 0, & \text{otherwise}. 
\end{cases}
\end{equation}
The model-dependent hyper-parameter $\epsilon \in [0.05, 0.15]$ is optimized on the validation set w.r.t the PAN 2021 metrics. 
The input of \gls{O2D2}, noted as $\boldsymbol{y}^{\text{O2D2}}$, is a concatenated vector of the \gls{LEVs}, i.e. $\big(\boldsymbol{y}_1 - \boldsymbol{y}_2 \big)^{\circ 2}$ and $\big(\boldsymbol{y}_1 + \boldsymbol{y}_2 \big)^{\circ 2}$, and the confusion matrix. This vector is fed into a three-layer architecture,
\begin{align}
\begin{split}
   \label{eq:o2d2pr}
    \boldsymbol{h}_1 
    &= \tanh\big(\boldsymbol{W}^{\textsc{O2D2}}_1 
    \boldsymbol{y}^{\textsc{O2D2}}
    + \boldsymbol{b}^{\textsc{O2D2}}_1\big),
\\
    \boldsymbol{h}_2
    &= \tanh\big(\boldsymbol{W}^{\textsc{O2D2}}_2 
    \boldsymbol{h}_1
    + \boldsymbol{b}^{\textsc{O2D2}}_2\big),
\\
    p_{\text{O2D2}}(\mathcal{H}_2| \boldsymbol{y}_1, \boldsymbol{y}_2)  
    &= \text{Sigmoid}\big(\boldsymbol{W}^{\textsc{O2D2}}_3 
    \boldsymbol{h}_2
    + \boldsymbol{b}^{\textsc{O2D2}}_3\big).
\end{split}
\end{align}
All trainable parameters are summarized in $\boldsymbol{\Gamma} = \big\{
\boldsymbol{W}^{\textsc{O2D2}}_i, \boldsymbol{b}^{\textsc{O2D2}}_i|i\in\{1,2,3\}
\big\}$.
The obtained prediction for hypothesis $\mathcal{H}_2$ is inserted into the cross-entropy loss, 
\begin{align}
 \begin{split}
\label{eq:losso2d2}
\loss^{\text{O2D2}}_{\boldsymbol{\Gamma}} 
    = l^{\text{O2D2}} \cdot \log \left\{ p_{\text{O2D2}}(\mathcal{H}_2|\boldsymbol{y}_1, \boldsymbol{y}_2) \right\}
     + (1-l^{\text{O2D2}}) \cdot \log \left\{ 1 - p_{\text{O2D2}}(\mathcal{H}_2|\boldsymbol{y}_1, \boldsymbol{y}_2) \right\}.
\end{split}
\end{align}

\begin{figure}[t]
\centering
\includegraphics[width=0.7\textwidth]{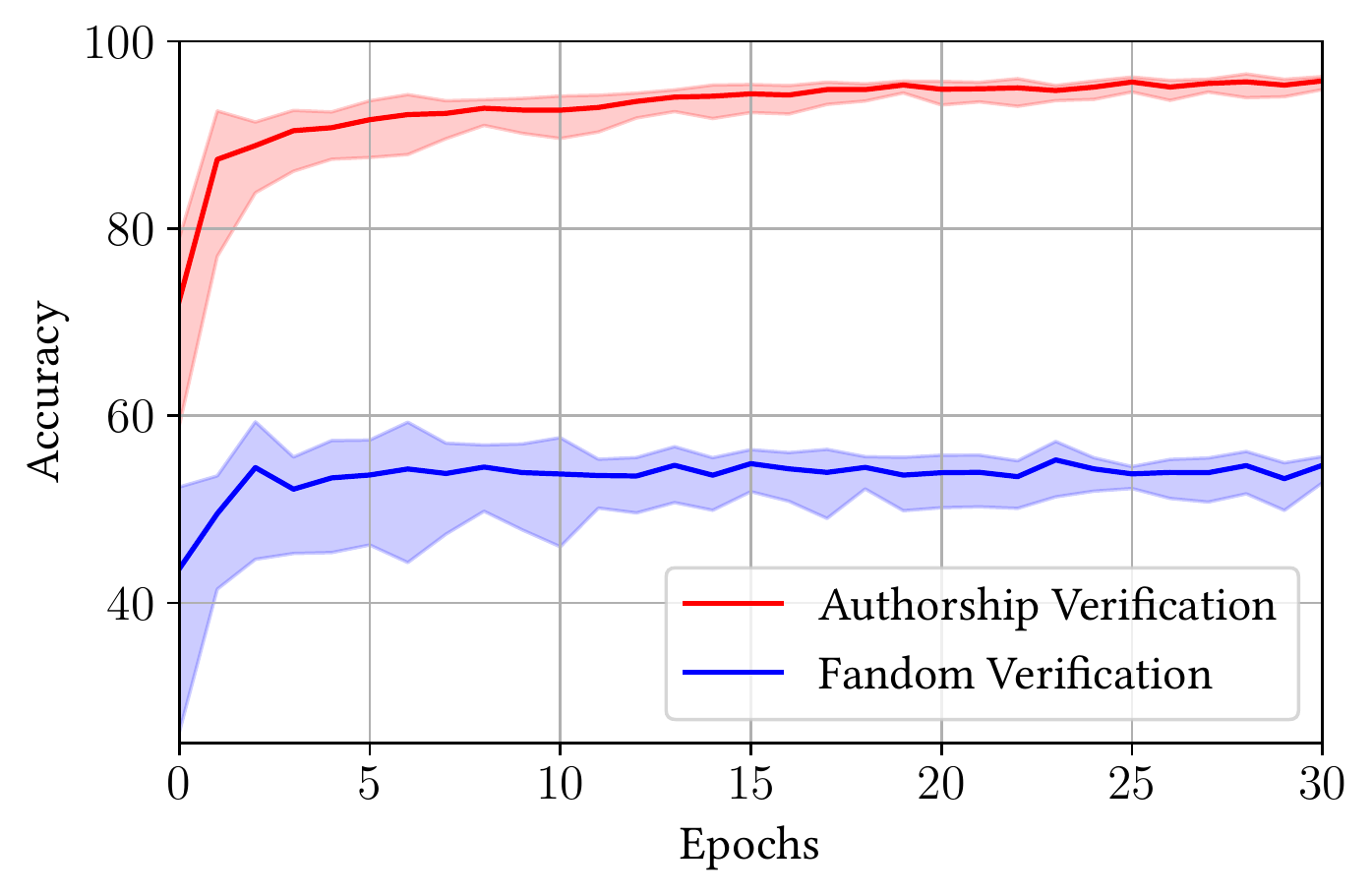}
\vspace*{-0.5cm}
\caption{Averaged accuracy curves (including mean and standard deviation) for the authorship and fandom verification outputs during training.} 
\label{fig:acc_f_a}
\end{figure}

\vspace*{-.2cm}
\subsection{Ensemble Inference}
\vspace*{-.1cm}
As a last step, an ensemble is constructed from $M$ trained models, $\mathcal{M}_1 , \ldots ,\mathcal{M}_M$, with $M$ being an odd number. 
Since all models are randomly initialized and trained on different re-sampled pairs in each epoch, we expect to obtain a slightly different set of weights/biases, which in turn produces different posteriors, especially for pairs under covariate shift.
We propose a majority voting for the non-responses. More precisely, the ensemble returns a non-response, if
\begin{align}
    \sum_{m=1}^{M}\mathbbm{1} \big[p_{\text{O2D2}}(\mathcal{H}_2|\boldsymbol{y}_1, \boldsymbol{y}_2,\mathcal{M}_m) \ge 0.5\big] >
    \left\lfloor\frac{M}{2}\right\rfloor,
\end{align}
where $\mathbbm{1}[\cdot]$ denotes the indicator function.
Otherwise, we define a subset of confident models,
$\mathcal{M}_c= \{\mathcal{M} | ~ p_{\text{O2D2}}(\mathcal{H}_2|\boldsymbol{y}_1, \boldsymbol{y}_2,\mathcal{M})
   < 0.5\}$, and return the averaged posteriors of its elements,
\begin{align}
    \mathbb{E}\big[ p_{\text{UAL}}(\mathcal{H}_1|\boldsymbol{y}_1, \boldsymbol{y}_2)
    \big]
    =
    \frac{1}{|\mathcal{M}_c|}\sum_{\mathcal{M} \in  \mathcal{M}_c}
   p_{\text{UAL}}(\mathcal{H}_1|\boldsymbol{y}_1, \boldsymbol{y}_2,\mathcal{M}).
\end{align}
Our submitted system consisted of an ensemble with $M=21$ trained models.

%% file: sections/experiments.tex
\section{Experiments}
\vspace*{-.2cm}
\label{sec:exp}

 \begin{table}[t]
 \centering
\caption{Averaged results (including mean and standard deviation) of the \gls{UAL} framework for different subset combinations on the calibration dataset.}
 \vspace*{-0.1cm}
\resizebox{0.9\textwidth}{!}{
\begin{tabular}{|c | c |c |c |c |c |c|}
    \hline
   \multicolumn{1}{|c|}{\multirow{2}{*}{\textbf{Subsets}}}
    & \multicolumn{6}{c|}{\textbf{PAN 2021 Evaluation Metrics}}
       \\ \cline{2-7}
    &\texttt{~AUC~}           &\texttt{~c@1~}       &\texttt{f\_05\_u}    &\texttt{~F1~}  &\texttt{~Brier~}      &\texttt{overall} 
    \\ \hline
 \multicolumn{1}{|c|}{\texttt{SA\_SF} + \texttt{DA\_DF}}
    &$99.8 \pm 0.0$  
    &$97.5 \pm 0.2$  
    &$97.2 \pm 0.3$  
    &$97.5 \pm 0.2$ 
    &$98.1 \pm 0.1$ 
    &$98.0 \pm 0.2$ 
        \\ 
 \multicolumn{1}{|c|}{\texttt{SA\_SF} + \texttt{DA\_SF}} 
    &$99.6 \pm 0.1$  
    &$95.9 \pm 0.4$  
    &$94.8 \pm 0.6$  
    &$96.0 \pm 0.4$
    &$97.1 \pm 0.2$ 
    &$96.7 \pm 0.4$ 
    \\
 \multicolumn{1}{|c|}{\texttt{SA\_DF} + \texttt{DA\_DF}}  
    &$98.1 \pm 0.1$  
    &$92.4 \pm 0.3$  
    &$94.8 \pm 0.3$  
    &$92.1 \pm 0.3$ 
    &$94.2 \pm 0.2$ 
    &$94.3 \pm 0.2$ 
    \\
 \multicolumn{1}{|c|}{\texttt{SA\_DF} + \texttt{DA\_SF}}  
    &$97.1 \pm 0.1$  
    &$90.9 \pm 0.3$  
    &$92.3 \pm 0.6$  
    &$90.6 \pm 0.3$ 
    &$93.1 \pm 0.2$ 
    &$92.8 \pm 0.3$ 
    \\ \hline
\end{tabular}
\label{tab:results_sub_pan}
}
\end{table}
 \begin{table}[t]
\caption{Averaged calibration results (including mean and standard deviation) of the \gls{UAL} framework for different subsets on the calibration dataset.}
\vspace*{-0.1cm}
 \centering
\resizebox{0.66\textwidth}{!}{
    \begin{tabular}{|c | c| c |c  |c|}
    \hline
   \multicolumn{1}{|c|}{\multirow{2}{*}{\textbf{Subsets}}}
    & \multicolumn{4}{c|}{\textbf{Calibration Metrics}}
       \\ \cline{2-5}
    &\texttt{~acc~}        &\texttt{~conf~}       &\texttt{~ECE~}    &\texttt{~MCE~} 
    \\ \hline
 \multicolumn{1}{|c|}{\texttt{SA\_SF} + \texttt{DA\_DF}}
    &$98.4 \pm 0.3$  
    &$97.4 \pm 1.0$  
    &$1.1 \pm 0.9$  
    &$5.7 \pm 3.6$ 
        \\ 
 \multicolumn{1}{|c|}{\texttt{SA\_SF} + \texttt{DA\_SF}} 
    &$96.1 \pm 0.6$  
    &$95.6 \pm 1.0$  
    &$1.3 \pm 0.8$  
    &$9.9 \pm 4.0$
    \\
 \multicolumn{1}{|c|}{\texttt{SA\_DF} + \texttt{DA\_DF}}  
    &$92.4 \pm 0.3$  
    &$93.4 \pm 1.2$  
    &$1.6 \pm 0.6$  
    &$6.2 \pm 2.5$ 
    \\
 \multicolumn{1}{|c|}{\texttt{SA\_DF} + \texttt{DA\_SF}}  
    &$90.9 \pm 0.3$  
    &$92.4 \pm 1.2$  
    &$2.0 \pm 0.7$  
    &$7.4 \pm 2.9$ 
    \\ \hline
\end{tabular}
\label{tab:results_sub_cal}
}
\end{table}

The PAN evaluation metrics and procedure are described in~\cite{kestemont:2020, kestemont:2021, potthast:2019n}. To capture the calibration capacity, we also provide the \gls{acc}, \gls{conf}, \gls{ECE} and \gls{MCE}~\cite{pmlr-v70-guo17a}.
All confidence values lie within the interval $[0.5, 1]$, since we are solving a binary classification task. Hence, to obtain confidence scores, the posterior values are transformed w.r.t.~the estimated authorship label, showing
$p(\mathcal{H}_1|\boldsymbol{y}_1, \boldsymbol{y}_2)$ if $\widehat{a}=1$ and $1 - p(\mathcal{H}_1|\boldsymbol{y}_1, \boldsymbol{y}_2)$ if $\widehat{a}=0$.
For both metrics, the confidence interval is discretized into a fixed number of bins. The \texttt{ECE} then reflects the average absolute error between confidence and accuracy of all bins, while the \texttt{MCE} returns the maximum absolute error. For \gls{acc} and \gls{conf}, we perform weighted macro-averaging w.r.t.~the number of trials in each bin.  

Inspired by the promising results in domain-adversarial training of neural networks in~\cite{journals/corr/GaninUAGLLML15, Bischoff2020}, we also experimented with an adversarial \textit{fandom verifier}: 
Starting with the document embeddings in Eq.~\eqref{eq:docemb}, we fed this vector into the author verification system (including \gls{DML}, \gls{BFS} and \gls{UAL}) and into an additional \textit{fandom verifier}, which is placed parallel to the author verification system. It has the same architecture but includes a gradient reversal layer and different trainable parameters.
However, in these experiments, we did not achieve any significant improvements by domain-adversarial training. 
Therefore, we independently optimized the fandom verifier by stopping the flow of the gradients from the fandom verifier to the authorship verification components, so that the training of the fandom verifier does not affect the target system at all.
Fig.~\ref{fig:acc_f_a} shows the obtained epoch-wise accuracies during training. It can be seen that the fandom accuracy stays around $55$\%, which indicates that the training strategy yields nearly topic-invariant stylometric representations, even without domain-adversarial training.

\vspace*{-.1cm}
\subsection{Results on the Calibration Dataset}
\vspace*{-.1cm}
We first evaluated the \gls{UAL} component on the calibration set (without non-responses) and 
calculated the respective PAN metrics for different combinations of the author-fandom subsets. Results are shown in Table~\ref{tab:results_sub_pan}.
To guarantee that the calculated metrics are not biased by an imbalanced dataset, we reduced the number of pairs to the smallest number of pairs of all subsets. 
Thus, all results in Table~\ref{tab:results_sub_pan} were computed from~\mbox{$2 \times 2,100$} pairs.
%
%
%
Unsurprisingly, best performance was obtained for the least challenging \texttt{SA\_SF} + \texttt{DA\_DF} pairs and the worst performance was seen for the most challenging \texttt{SA\_DF} + \texttt{DA\_SF} pairs. 
We continued to optimize our system w.r.t~this most challenging subset combination in particular, 
even though we specifically expect to see \texttt{SA\_DF} + \texttt{DA\_DF} pairs in the PAN 2021 evaluation set.

Table~\ref{tab:results_sub_cal} additionally provides the corresponding calibration metrics. 
Analogously to the PAN metrics, the \texttt{ECE} 
consistently increases from the least to the most challenging data scenarios.
Interestingly, our system is \textit{under-confident} for \texttt{SA\_SF} pairs, i.e.~\texttt{conf} $<$ \texttt{acc}. 
The predictions then change to be \textit{over-confident} (\texttt{conf} $>$ \texttt{acc}) for \texttt{SA\_DF} pairs.

 \begin{table}[t]
\caption{Results for PAN 2021 evaluation metrics on the validation datset.}
 \vspace*{-0.1cm}
 \centering
\resizebox{0.9\textwidth}{!}{
\begin{tabular}{|c c | c |c |c |c |c |c|}
    \hline
   \multicolumn{2}{|c|}{\multirow{2}{*}{\textbf{Model}}}
    & \multicolumn{6}{c|}{\textbf{PAN 2021 Evaluation Metrics}}
       \\ \cline{3-8}
    & &\texttt{~AUC~}           &\texttt{~c@1~}       &\texttt{f\_05\_u}    &\texttt{~F1~}  &\texttt{~Brier~}      &\texttt{overall} 
    \\ \hline
    \multirow{4}{*}{\begin{adjustbox}{angle=90}\textbf{single}\end{adjustbox}} 
    & \multicolumn{1}{|c|}{\textsc{DML}}
    &$97.2 \pm 0.1$  
    &$91.3 \pm 0.3$  
    &$90.5 \pm 0.6$  
    &$89.6 \pm 0.4$ 
    &$93.2 \pm 0.4$ 
    &$92.4 \pm 0.2$ 
        \\ 
    & \multicolumn{1}{|c|}{BFS} 
    &$97.1 \pm 0.1$  
    &$91.0 \pm 0.3$  
    &$90.7 \pm 0.8$  
    &$89.2 \pm 0.5$
    &$93.2 \pm 0.1$ 
    &$92.3 \pm 0.2$ 
    \\
    & \multicolumn{1}{|c|}{UAL}  
    &$97.2 \pm 0.1$  
    &$91.3 \pm 0.3$  
    &$90.7 \pm 0.5$  
    &$89.6 \pm 0.4$ 
    &$93.5 \pm 0.2$ 
    &$92.5 \pm 0.2$ 
    \\
    & \multicolumn{1}{|c|}{O2D2}  
    &$97.1 \pm 0.1$  
    &$93.8 \pm 0.2$  
    &$88.1 \pm 0.6$  
    &$93.5 \pm 0.3$ 
    &$93.4 \pm 0.1$ 
    &$93.2 \pm 0.2$ 
    \\ \hline
    \multicolumn{2}{|c|}{ensemble}  
    &$97.8 $  
    &$92.5 $  
    &$92.1 $  
    &$90.9 $ 
    &$94.3 $ 
    &$93.5 $ 
    \\ \hline
    \multicolumn{2}{|c|}{ensemble + O2D2}  
    &$97.7 $  
    &$94.8 $  
    &$90.0 $  
    &$94.5 $ 
    &$94.2 $ 
    &$94.2 $ 
    \\ \hline
\end{tabular}
\label{tab:results_pan_val}
}
\end{table}

 \begin{table}[t]
 \caption{Results for the calibration metrics on the validation dataset.}
 \vspace*{-0.1cm}
 \centering
\resizebox{0.66\textwidth}{!}{
\begin{tabular}{|c c | c| c |c  |c|}
    \hline
   \multicolumn{2}{|c|}{\multirow{2}{*}{\textbf{Model}}}
    & \multicolumn{4}{c|}{\textbf{Calibration Metrics}}
       \\ \cline{3-6}
    &   &\texttt{~acc~}        &\texttt{~conf~}       &\texttt{~ECE~}    &\texttt{~MCE~} 
    \\ \hline
    \multirow{4}{*}{\begin{adjustbox}{angle=90}\textbf{single}\end{adjustbox}} 
    & \multicolumn{1}{|c|}{DML}
    &$91.3 \pm 0.3$  
    &$87.9 \pm 2.7$  
    &$3.4 \pm 2.7$  
    &$~9.0 \pm 3.6$ 
        \\ 
    & \multicolumn{1}{|c|}{BFS} 
    &$91.0 \pm 0.3$  
    &$90.0 \pm 2.3$  
    &$2.3 \pm 1.5$  
    &$~6.2 \pm 3.0$
    \\
    & \multicolumn{1}{|c|}{UAL}  
    &$91.3 \pm 0.3$  
    &$92.3 \pm 1.2$  
    &$1.6 \pm 0.6$  
    &$~5.8 \pm 2.2$ 
    \\
    & \multicolumn{1}{|c|}{O2D2}  
    &$91.4 \pm 0.3$  
    &$90.9 \pm 1.1$  
    &$2.3 \pm 0.5$  
    &$10.7 \pm 2.7$ 
    \\ \hline
    \multicolumn{2}{|c|}{ensemble}  
    &$92.5$
    &$91.2$  
    &$1.2 $  
    &$2.9$
    \\ \hline
    \multicolumn{2}{|c|}{ensemble + O2D2}
    &$92.6$  
    &$91.8$  
    &$1.5 $  
    &$10.2$
    \\ \hline
\end{tabular}
\label{tab:results_cal_val}
}
\end{table}
\begin{figure}[t]
\centering
\begin{subfigure}[t]{0.45\textwidth}
\centering
\includegraphics[width=0.99\textwidth]{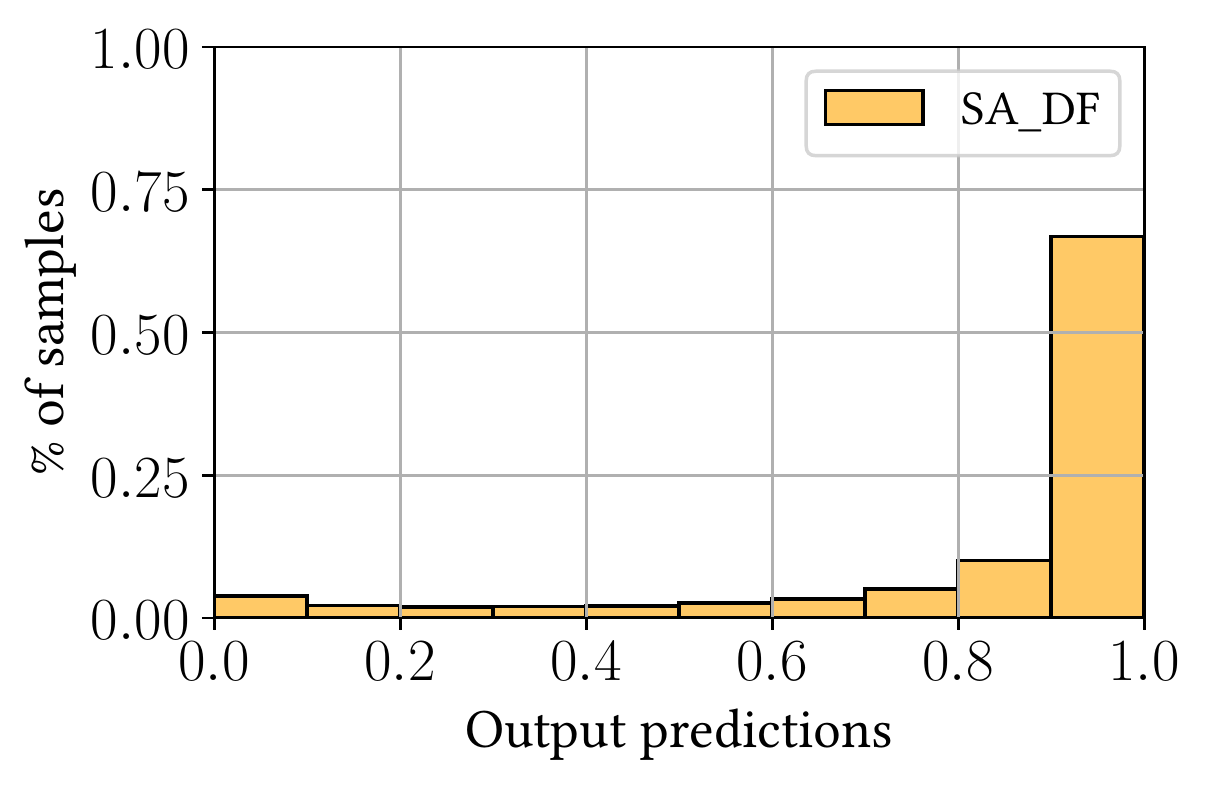}
\caption{\texttt{SA\_DF} pairs without O2D2.}
\end{subfigure}
\quad
\begin{subfigure}[t]{0.45\textwidth}
\centering
\includegraphics[width=0.99\textwidth]{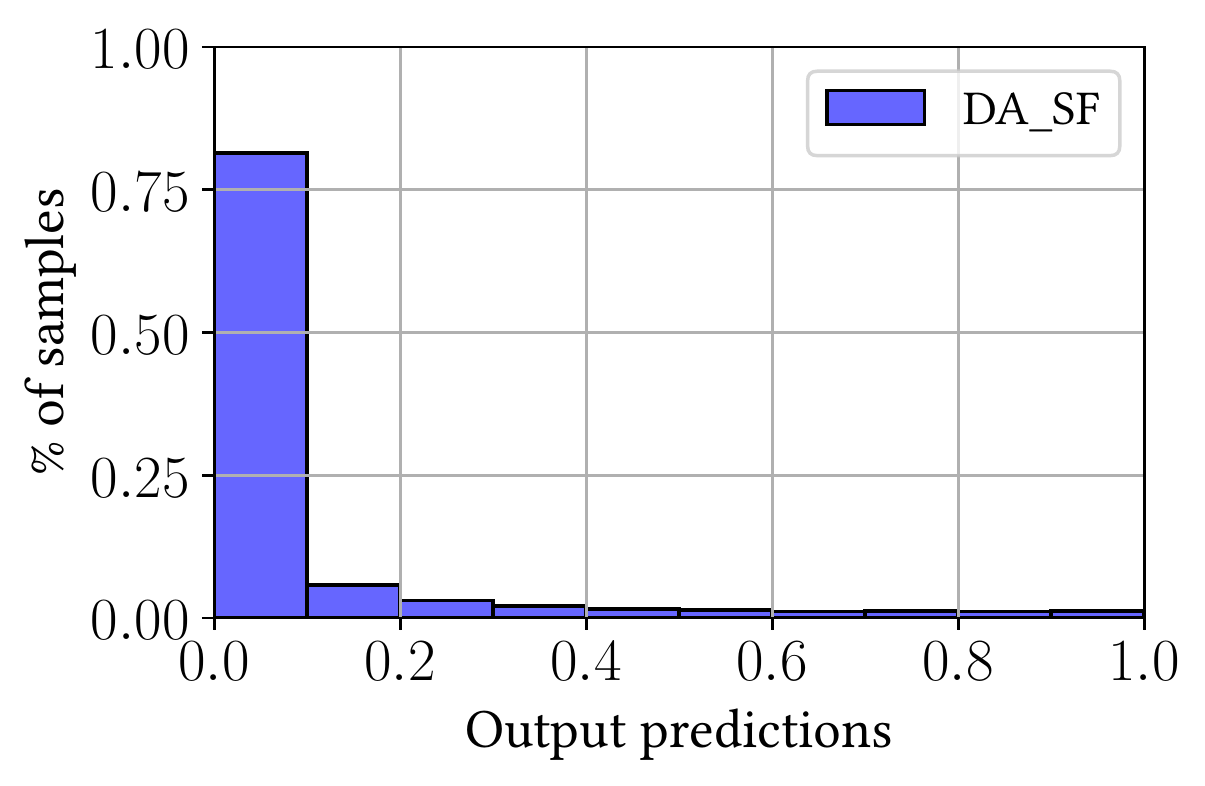}
\caption{\texttt{DA\_SF} pairs without O2D2.}
\end{subfigure}
\begin{subfigure}[t]{0.45\textwidth}
\centering
\includegraphics[width=0.99\textwidth]{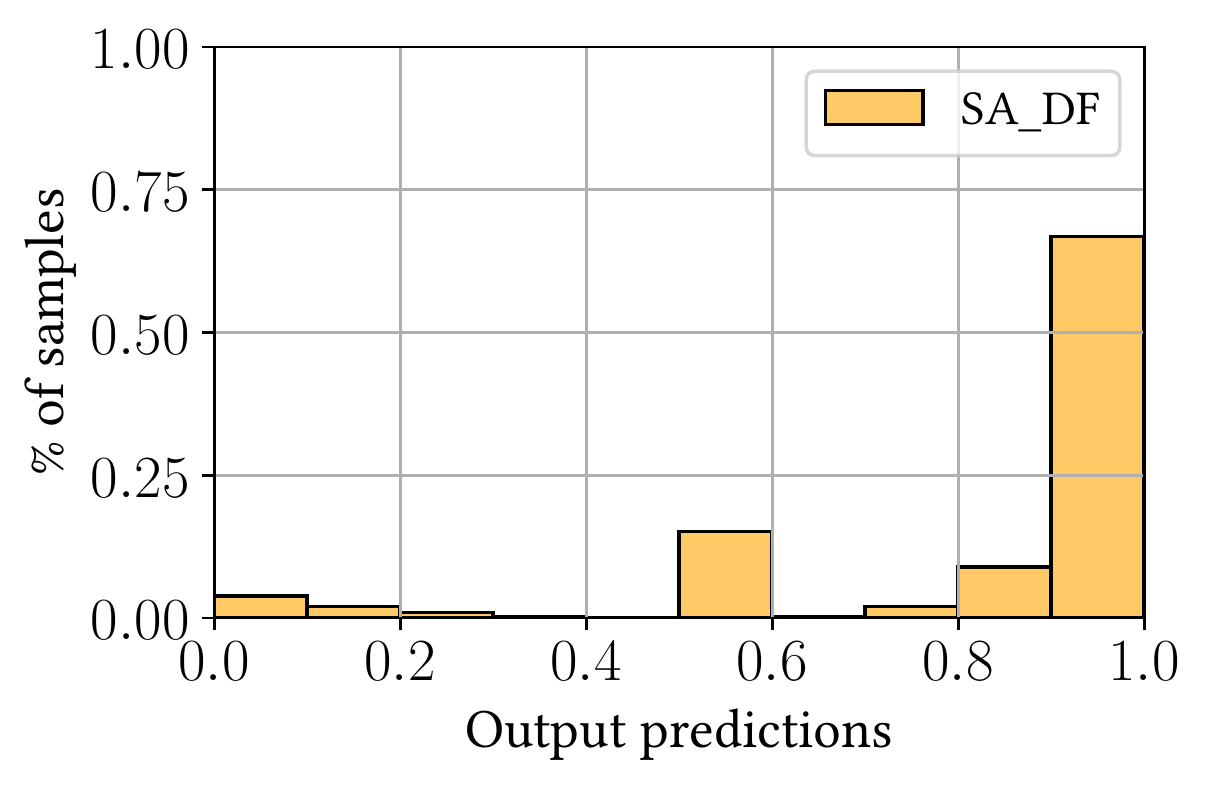}
\caption{\texttt{SA\_DF} pairs including O2D2.}
\end{subfigure}
\quad
\begin{subfigure}[t]{0.45\textwidth}
\centering
\includegraphics[width=0.99\textwidth]{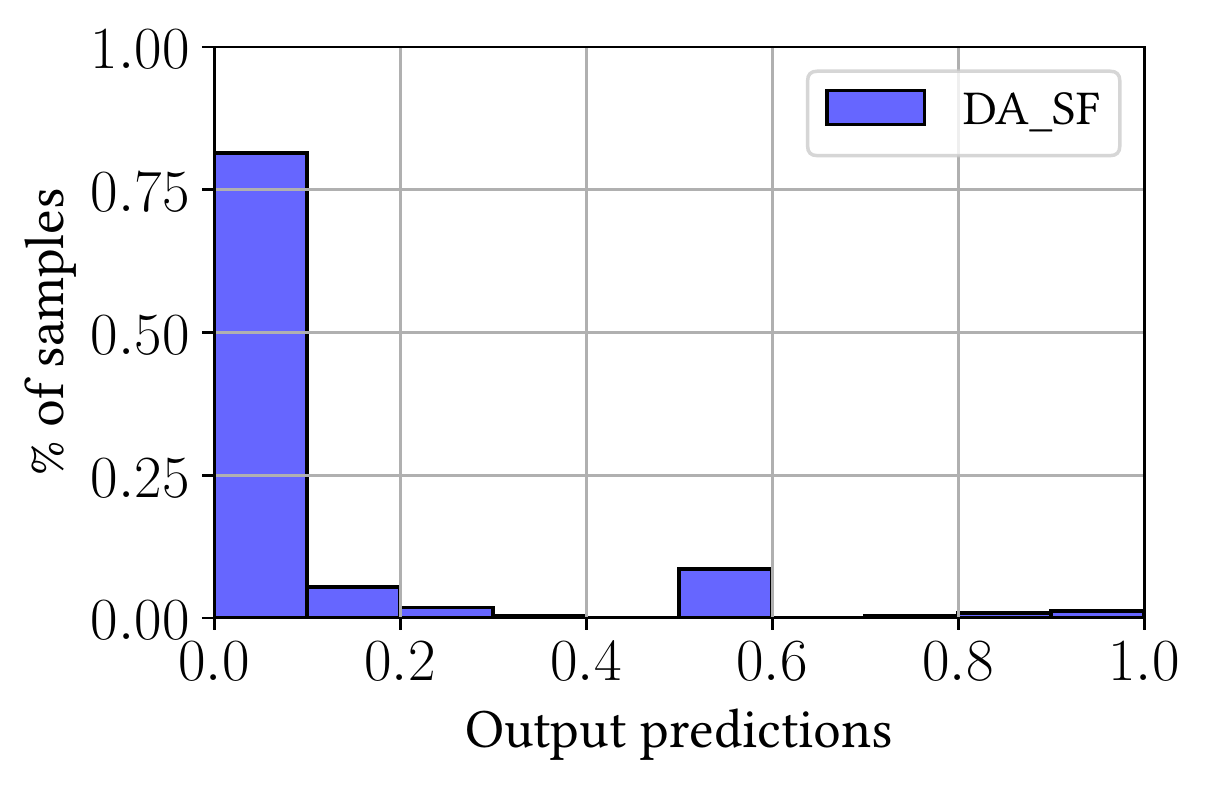}
\caption{\texttt{DA\_SF} pairs including O2D2.}
\end{subfigure}
\vspace*{-0.2cm}
\caption{Posterior histograms on the validation dataset.} 
\label{fig:cal_his_all_val}
\end{figure}

\vspace*{-.2cm}
\subsection{Results on the Validation Dataset}
\vspace*{-.1cm}
Next, we separately provide experimental results for all system components on the validation dataset, since \gls{O2D2} has been trained on the calibration dataset. 
The first four rows in Tables~\ref{tab:results_pan_val} and~\ref{tab:results_cal_val} summarize the PAN metrics and the corresponding calibration measures averaged over all ensembles models.

The overall score of the \gls{UAL} component in the third row of Table~\ref{tab:results_pan_val} is on par with the \gls{DML} and \gls{BFS} components and slightly lower compared to the corresponding \gls{UAL} score measured on the calibration dataset in Table~\ref{tab:results_sub_pan}. Nevertheless, we do not observe significant differences in the metrics for both datasets, which shows the robustness and generalization of our system.

Going from the third to the fourth row in Table~\ref{tab:results_pan_val},
it can be observed that the overall score, boosted by \texttt{c@1} and \texttt{F1}, significantly increases from $92.5$ to $93.2$. Hence, the model performs better if we take undecidable trials into account. However, the \texttt{f\_05\_u} score decreases, since it treats non-responses as false negatives. The percentage of undecidable trials generally ranges from 8\% to 11\%.

In Table~\ref{tab:results_cal_val}, we see that both, the \gls{BFS} and \gls{UAL} components 
notably improve the \texttt{ECE} and \texttt{MCE} metrics. However, an insertion of non-responses via O2D2 significantly increases the \gls{MCE}. This can be explained by the posterior histograms in Fig.~\ref{fig:cal_his_all_val}. 
The plots (a) and (b) show the histograms for \texttt{SA\_DF} and \texttt{DA\_SF} pairs without applying \gls{O2D2} to define non-responses. In contrast, plots (c) and (d) present the corresponding histograms including the $0.5$-values of non-responses.
The effect of \gls{O2D2} is that most of the trials, whose posteriors fall within the interval $[0.3, 0.7]$, are 
eventually declared as undecidable. Hence, the system correctly predicts nearly all of the remaining as confidently assigned trials around $0.7$/$0.8$ for same-author pairs or $0.2$/$0.3$ for different-author pairs. As a result, we see a large gap (i.e. \texttt{conf} $<<$ \texttt{acc}) between the confidence score and the averaged accuracy in these bins. 

The last two rows in Tables~\ref{tab:results_pan_val} and~\ref{tab:results_cal_val} show the performance of the ensemble, first without and then with non-responses, to show the effect of \gls{O2D2}. On the validation set, our ensemble with \gls{O2D2} returns non-responses in $9$\% of the test cases. Comparing the last two rows, we obtain the highest overall score with our proposed framework, which ultimately presents our final submission. 

\vspace*{-.2cm}
\subsection{Results on the PAN 2021 Evaluation Dataset}
\vspace*{-.1cm}
To conclude this section, we present our results on the official PAN 2021 evaluation set. 
The performance for both, the early-bird and the final submission, can be found in Table~\ref{tab:resultsfinal}. We also provide the reported result on the PAN 2020 evaluation set for the predecessor model. 

Unsurprisingly, the early-bird overall score (single model) on the PAN 2021 evaluation set is slightly higher, since it contains \texttt{DA\_DF} pairs instead of \texttt{DA\_SF} pairs. 
The main difference is, unexpectedly, given by the \texttt{f\_05\_u} score, which increases from $89.3$\% to $94.6$\%. In our opinion, this is caused by returning a lower number of non-responses, which would also explain the lower values for
\texttt{c@1} and \texttt{F1}.

Comparing the early-bird ($2^{nd}$ row) with the final submission ($4^{th}$ row), we can further significantly increase the overall score by $1.5\%$. We assume that the ensemble now returns a higher number of non-responses, which results in a slightly lower \texttt{f\_05\_u} score. Conversely, we can observe improved values for the \texttt{c@1}, \texttt{F1} and \texttt{brier} scores.

The last row displays the achieved PAN 2020 results. As can be seen, our final submission ends up with a higher overall score (plus $2\%$) by significantly improving all single metrics, although the PAN competition moved from a closed-set to open-set shared task, illustrating the efficiency of the proposed extensions.

 \begin{table}[t]
\caption{Results of the early-bird (first two rows) and the final submission runs.}
\vspace*{-0.1cm}
\centering
\resizebox{1.0\textwidth}{!}{
  \begin{tabular}{|r |c | c |c |c |c |c | c|}
    \hline
 Dataset & Model type
 &\texttt{~AUC~}           &\texttt{~c@1~}       &\texttt{~f\_05\_u~}    &\texttt{~F1~} &\texttt{~brier~}     &\texttt{overall} \\ \hline
Validation dataset &single-21
    &$97.2$  
    &$93.6$  
    &$89.3$  
    &$92.9$ 
    &$93.3$ 
    &$93.3$ 
    \\
PAN 21 evaluation dataset  &single-21
&$98.3$      &$92.6$    &$94.6$   &$92.1$      &$92.7$   &$94.0$
\\ \hline
Validation dataset  &ensemble-21
    &$97.7$  
    &$94.8$  
    &$90.1$  
    &$94.4$ 
    &$94.2$ 
    &$94.2$
 \\
PAN 21 evaluation dataset &ensemble-21
&$98.7$      &$95.0$    &$93.8$ &$95.2$         &$94.5$   &$95.5$
    \\ \hline
PAN 20 evaluation dataset &ensemble-20
&$96.9$      &$92.8$    &$90.7$   &$93.6$      &-   &$93.5$
    \\ \hline
\end{tabular}
\label{tab:resultsfinal}
}
\end{table}

%% file: sections/conclusion.tex
\vspace*{-.3cm}
\section{Conclusion}
\vspace*{-.2cm}
In this work, we presented \gls{O2D2}, which captures undecidable trials and supports our hybrid neural-probabilistic end-to-end framework for authorship verification. 
We made use of the early-bird submission to receive a preliminary assessment of how the framework behaves on the novel open-set evaluation. Finally, based on the presented results, we submitted an \gls{O2D2}-supported ensemble to the shared task, which clearly outperformed our own system from 2020 as well as the new submissions to the PAN 2021 \gls{AV} task.

These results support our hypothesis that modeling aleatoric and epistemic uncertainty and using them for decision support is a beneficial strategy---not just for responsible ML, which needs to be aware of the reliability of its proposed decisions, but also, importantly, for achieving optimal performance in real-life settings, where distributional shift is almost always hard to avoid.
